\documentclass[letterpaper, 10 pt, conference]{ieeeconf}  

\IEEEoverridecommandlockouts                              

\usepackage{graphics} 
\usepackage{epsfig} 
\usepackage{mathptmx} 
\usepackage{times} 
\usepackage{amsmath} 
\usepackage{amssymb}  
\usepackage{cuted}
\usepackage{caption}
\usepackage{adjustbox}
\usepackage{booktabs}
\usepackage{multirow}
\usepackage{float} 
\title{\LARGE \bf
Trajectory Conditioned Cross-embodiment Skill Transfer
}

\author{%
  YuHang Tang$^{1,2}$\quad Yixuan Lou$^1$\quad Pengfei HAN$^1$\quad Haoming Song$^{2,3}$ \\
  Xinyi Ye$^2$\quad Dong Wang$^2$\quad Bin Zhao$^{\dagger1,2}$\\
  $^1$Northwestern Polytechnical University \quad
  $^2$Shanghai AI Laboratory
  $^3$Shanghai Jiao Tong University
}

\begin{document}

\maketitle
\begin{strip}
\vspace{-1.5cm}
\begin{center}
\includegraphics[width=1.0\textwidth]{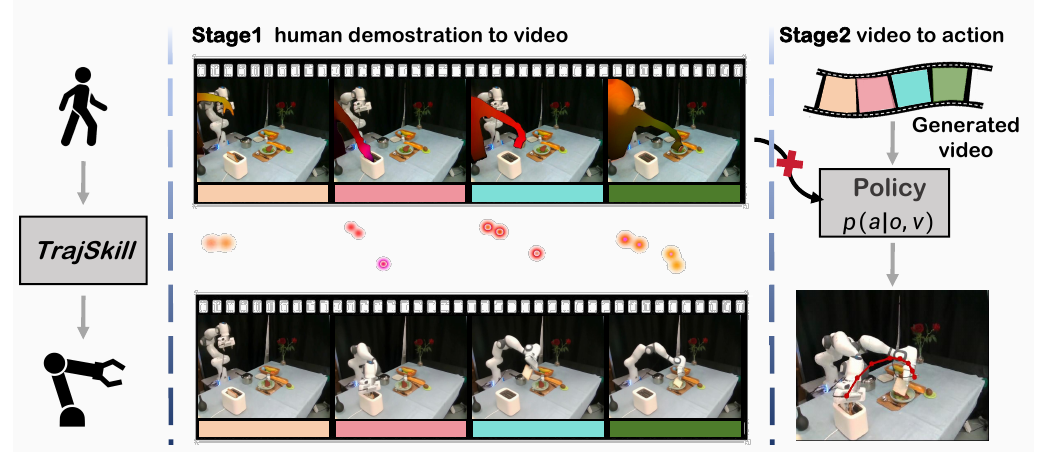}
    \captionof{figure}{Overview of the TrajSkill from human to robot action. TrajSkill leverages sparse optical flow as a universal motion representation, achieving zero-shot imitation without reinforcement learning or paired datasets. In Stage 1, dense optical flow is extracted from human demonstrations and sampled into sparse optical flow to guide video generation. In Stage 2, the generated video is translated into robot actions using a learned policy, enabling the robot to mimic the demonstrated task.}
    \label{fig:sim_fig2}
\end{center}
\vspace{-0.5cm}
\end{strip}

\thispagestyle{empty}
\pagestyle{empty}


\begin{abstract}

Learning manipulation skills from human demonstration videos presents a promising yet challenging problem, primarily due to the significant embodiment gap between human body and robot manipulators. Existing methods rely on paired datasets or hand-crafted rewards, which limit scalability and generalization.  
We propose \textbf{TrajSkill}, a framework for \textit{Trajectory Conditioned Cross-embodiment Skill Transfer}, enabling robots to acquire manipulation skills directly from human demonstration videos. Our key insight is to represent human motions as \emph{sparse optical flow trajectories}, which serve as embodiment-agnostic motion cues by removing morphological variations while preserving essential dynamics. Conditioned on these trajectories together with visual and textual inputs, TrajSkill jointly synthesizes temporally consistent robot manipulation videos and translates them into executable actions, thereby achieving cross-embodiment skill transfer.  
Extensive experiments are conducted, and the results on simulation data (MetaWorld) show that TrajSkill reduces FVD by 39.6\% and KVD by 36.6\% compared with the state-of-the-art, and improves cross-embodiment success rate by up to 16.7\%. Real-robot experiments in kitchen manipulation tasks further validate the effectiveness of our approach, demonstrating practical human-to-robot skill transfer across embodiments.

\end{abstract}

\section{Introduction}

Robotic manipulation learning from human demonstration videos has been a long, compelling yet challenging task in embodied intelligence. While Human videos naturally capture manipulation dynamics, the direct transfer of these skills remains impracticable due to substantial differences in morphology, kinematic constraints, and embodiment between the human body and robot manipulators. Previous approaches have attempted to bridge this gap through reinforcement learning with hand-crafted reward functions~\cite{liu2018imitation,shao2021concept2robot}, meta-learning for one-shot imitation~\cite{yu2018one}, or domain alignment techniques between human and robot embodiments~\cite{majumdar2023we}. However, these methods often depend on costly human interventions, paired datasets, or brittle alignment strategies, which limits their scalability and practical deployment in real-world scenarios.

Recent advances in video generation models open new avenues for robot policy learning~\cite{bao2024vidu}, offering the potential to synthesize long-horizon motion sequences that can inform planning~\cite{brooks2024video,yang2024cogvideox}. 
Parallel to these developments, a growing body of work has sought to directly learn robot policies from human demonstration videos. 
For example, Learning by Watching~\cite{xiong2021learning} translates human motions into robot actions via keypoint-based representations, but relies on accurate mappings between embodiments. 
Vid2Robot~\cite{jain2024vid2robot} introduces an end-to-end video-conditioned policy that learns from paired human videos and robot trajectories, yet its dependence on large paired datasets and embodiment alignment remains a bottleneck. 
More recently, Human2Robot~\cite{xie2025human2robot} formulates video-to-action transfer as a diffusion-based generative task on a large paired dataset, while Motion Tracks~\cite{ren2025motion} introduces a keypoint retargeting network for few-shot transfer; both approaches, however, remain sensitive to embodiment discrepancies. 
Despite these advances, existing methods typically require paired datasets, hand-crafted rewards, or explicit human–robot alignment strategies, all of which stem from the fundamental embodiment gap between human and robot morphologies. This reliance limits their scalability and hinders the development of morphology-invariant representations for robust skill transfer.

To solve aforementioned challenges, it is essential to discover a motion representation that is compact, embodiment-agnostic, and retains essential task dynamics. We observe that employing sparse optical flow effectively filters out appearance and morphological differences while preserving the key motion intent, thereby achieving embodiment invariance. Building on this insight, we propose \textbf{TrajSkill}, a trajectory conditioned cross-embodiment skill transfer framework. TrajSkill leverages \emph{sparse optical flow trajectories} extracted from human demonstrations as a unified motion representation, which eliminates embodiment-specific appearance while preserving dynamic motion patterns. Conditioned on these trajectories, TrajSkill converts human demonstrations into executable robot policies, achieving zero-shot imitation without reinforcement learning or paired datasets.

Our contributions are summarized as follows:
\begin{itemize}

\item We propose {TrajSkill}, a framework for {trajectory conditioned cross-embodiment skill transfer}, which jointly enables controllable video generation and executable robot policy learning directly from human demonstration videos.
\item We introduce sparse optical flow trajectories as an embodiment-agnostic representation that bridges the morphological gap between human and robot embodiments, providing effective motion cues for skill transfer.

\item We validate {TrajSkill} extensively across a diverse set of manipulation benchmarks encompassing dozens of tasks, demonstrating consistent improvements in video generation quality, cross-embodiment success rates, and real-robot skill execution in challenging kitchen manipulation scenarios.

\end{itemize}

\section{Related Work}

\subsection{Video Diffusion Models}
Video Diffusion Models (VDMs) have recently achieved impressive progress in generating high-quality video content. Early methods extended image diffusion architectures by adding temporal convolutions and attention layers within a UNet backbone~\cite{ho2022imagen,ho2022video}. While these approaches demonstrated the feasibility of diffusion-based video synthesis, their scalability and long-horizon consistency were fundamentally constrained. Subsequent works such as VideoCrafter~\cite{chen2023videocrafter1} and Stable Video Diffusion~\cite{blattmann2023stable} expanded training to larger datasets but still struggled with generating temporally coherent long sequences.

The introduction of Diffusion Transformers (DiT) marked a paradigm shift, enabling more scalable and unified sequence modeling~\cite{polyak2024movie}. Large-scale systems such as Sora~\cite{brooks2024video}, Vidu~\cite{bao2024vidu}, and CogVideoX~\cite{yang2024cogvideox} demonstrate the capability of DiT to generate high-definition videos extending to tens of seconds or more, with flexible aspect ratios and improved motion fidelity. Building upon these advances, our work adopts a DiT backbone for trajectory conditioned video synthesis with enhanced temporal coherence.

\subsection{Motion Control in Video Generation and Robotic Manipulation}
Beyond generating realistic appearance, controllable motion generation is critical for both video synthesis and robotics. In video generation, prior works have proposed conditioning on reference videos~\cite{liumotiondirector,jeong2024vmc}, structural cues such as depth maps or sketches~\cite{wang2023videocomposer}, or object masks~\cite{dai2023fine,wu2024draganything}. Recently, trajectory-based conditioning has attracted attention due to its physical intuitiveness, allowing users to directly specify object or camera motion~\cite{yin2023dragnuwa,wang2024motionctrl}. However, these methods often struggle with motion consistency over long horizons.

In robotics, generative video models have been explored as policy representations, where predicted video plans are mapped into executable actions. Works such as UniSim~\cite{yang2023learning} and UniPi~\cite{du2023learning} use text- or image-conditioned video prediction for robot interaction planning, while SuSIE~\cite{black2023zero} and AVDC~\cite{ko2023learning} incorporate autoregressive or custom diffusion architectures to infer actions from predicted trajectories or optical flow. More recent approaches such as SEER~\cite{gu2023seer} and This\&That~\cite{wang2024language} leverage language and multimodal signals for temporally aligned video generation. These works underscore the potential of video diffusion for scalable policy learning, yet they leave unresolved the challenge of generating robot-consistent motion videos with high fidelity and controllable trajectories.

\subsection{Cross-embodiment Learning from Human Videos}
Learning from human demonstrations offers a scalable alternative to costly robot-collected data. Prior methods construct rewards from human videos~\cite{liu2018imitation,shao2021concept2robot}, perform one-shot imitation via meta-learning~\cite{yu2018one}, or learn aligned visual embeddings across human and robot domains~\cite{nair2022r3m}. Others extract affordance cues~\cite{liu2022joint}, human-object interactions~\cite{goyal2022human}, or explicit hand trajectories and keypoints~\cite{wang2023mimicplay,pavlakos2024reconstructing}. Despite recent progress, these approaches often rely on reinforcement learning loops, require paired datasets, or suffer from brittle trajectory retargeting due to morphological mismatches.

To address these challenges, we introduce sparse optical flow trajectories as an embodiment-agnostic motion representation. By projecting both human and robot motions into a unified 2D trajectory space and conditioning a Diffusion Transformer-based video generator on these signals, we achieve cross-embodiment skill transfer without reinforcement learning optimization or paired training data, enabling scalable one-shot imitation from human demonstrations.

\section{Method}

Our framework realizes {trajectory conditioned cross-embodiment skill transfer} by leveraging sparse optical flow as an embodiment-agnostic representation. TrajSkill consists of three components: (1) embodiment-invariant flow sampling, (2) trajectory conditioned robot execution, and (3) cross-embodiment skill transfer. 

\begin{figure}[t]
    \centering
    \includegraphics[width=0.875\linewidth]{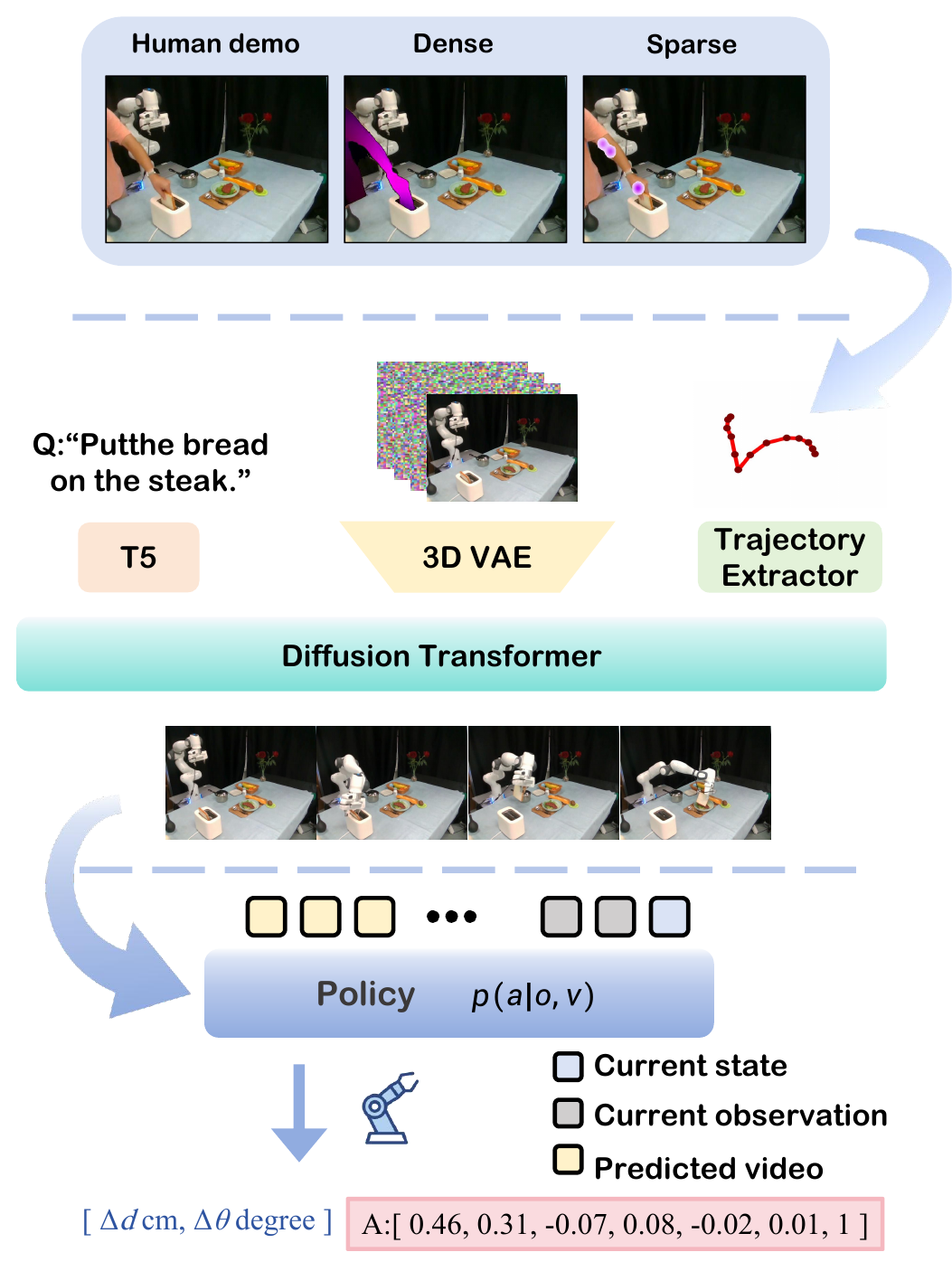}
    \caption{Unified illustration of the TrajSkill framework. 
    \textbf{Top:} Embodiment-Invariant Flow Sampling. From a human demonstration video frame (left), dense optical flow is computed by RAFT~\cite{teed2020raft} (middle), and sparse keypoint trajectories are sampled according to the flow magnitude and propagated over time (right). 
    \textbf{Middle and Bottom:} Overview of the Trajectory Conditioned Robot Execution. Given a task description, the T5 model interprets the instruction, a 3D VAE extracts spatial features, and the trajectory extractor provides sparse flow signals. These are fused within a Diffusion Transformer to predict robot motion videos, which are then decoded by the policy \(p(a|o,v)\) into executable actions.}
    \label{fig:arc}
    \vspace{-10pt}
\end{figure}

\subsection{Embodiment-Invariant Flow Sampling}
As shown in the top of Fig.~\ref{fig:arc}, our {Embodiment-Invariant Flow Sampling} computes dense optical flow and reduces it to compact sparse trajectories. Given a human demonstration video, we first compute the dense optical flow \(\mathbf{F}_t \in \mathbb{R}^{H \times W \times 2}\) between consecutive frames using RAFT~\cite{teed2020raft}. The flow \(\mathbf{F}_t\) represents the displacement between pixel locations across frames. To construct a compact trajectory representation, we define a grid of candidate positions with a stride \(\lambda\):
\begin{equation}
C = \left\{ (x, y) \mid x \in \{o_w, o_w + \lambda, \dots, W\}, y \in \{o_h, o_h + \lambda, \dots, H\} \right\},
\end{equation}
where \(o_w\) and \(o_h\) are random offsets within the image dimensions \(H \times W\). Each candidate is sampled with probability proportional to its initial flow magnitude. Specifically, the probability for a candidate \((x, y)\) is:
\begin{equation}
p_{(x, y)} = \frac{\|\mathbf{F}_0(x, y)\|_2}{\sum_{(x', y') \in C} \|\mathbf{F}_0(x', y')\|_2},
\end{equation}
where \(\|\mathbf{F}_0(x, y)\|_2\) is the \(L_2\)-norm of the flow vector at position \((x, y)\) in the first frame. The probabilistic sampling ensures that regions with stronger motion are more likely to be selected as candidate positions, effectively focusing on areas with significant movement.

To determine the actual sampled keypoints, we first draw the number of samples \(N\) uniformly from a maximum budget \(N_{\max}\):
\begin{equation}
N \sim \mathrm{Uniform}\{1,\dots,N_{\max}\},
\end{equation}
and then select \(N\) distinct candidates without replacement according to the probability distribution \(p_{(x,y)}\). The initial keypoint set:
\begin{equation}
K_0 = \{(x_{u_1},y_{u_1}), \dots, (x_{u_N},y_{u_N})\}, \quad (x_{u_k},y_{u_k}) \sim p_{(x,y)}.
\end{equation}

Next, we propagate the selected keypoints through time by integrating the local flow vectors. At each timestep \(t\), the new position \((x^{t+1}, y^{t+1})\) of a keypoint is updated using the flow vector at current position \((x^t, y^t)\):
\begin{equation}
(x^{t+1}, y^{t+1}) = (x^t, y^t) + \mathbf{F}_t(x^t, y^t),
\end{equation}
where \(\mathbf{F}_t(x^t, y^t)\) is the flow vector at position \((x^t, y^t)\) at time \(t\). By repeating this process, we obtain a set of sparse trajectories \textit{T}:
\begin{equation}
T = \{(x_i^t, y_i^t)\}_{i, t},
\end{equation}
where each trajectory \(T_i = \{(x_i^t, y_i^t)\}_t\) represents the path of a selected keypoint over time.

Finally, to mitigate noise and improve spatial consistency, we apply a Gaussian blur smoothing to the sparse flow field before usage. Concretely, each flow channel is convolved with a normalized isotropic Gaussian kernel:
\begin{equation}
\tilde{S}^{(d)}_t(u,v) = \sum_{i=-k}^{k} \sum_{j=-k}^{k} G(i,j)\, S^{(d)}_t(u-i,v-j),
\end{equation}
where \(S^{(d)}_t\) denotes the sparse flow channel \(d \in \{x,y\}\), \(G(i,j) \) is the discrete Gaussian kernel. The smoothed flow \(\tilde{S}_t=(\tilde{S}^{(x)}_t,\tilde{S}^{(y)}_t)\) is then used to construct the final trajectory representation.
It discards the embodiment-specific appearance details, focusing solely on the essential motion intent, which is key for analyzing the human demonstration while maintaining compactness and efficiency in trajectory representation. In this way, the embodiment gap is eliminated, ensuring that the representation consistently reflects task dynamics rather than morphological variations.

\subsection{Trajectory Conditioned Robot Execution} 

\textbf{Trajectory Conditioned Video Generation} 
To synthesize robot manipulation sequences conditioned on  instruction and trajectory inputs, we adopt a latent diffusion transformer (DiT) backbone~\cite{yang2024cogvideox}. Unlike prior UNet-based video diffusion models that rely on local convolutional receptive fields, our architecture leverages global attention to capture long-range temporal dependencies, thereby enabling scalable modeling of long-horizon robot motion with improved temporal coherence. As shown in the middle of Fig.~\ref{fig:arc}, the process is defined as:
\begin{equation} V_r = {G}(I_0, C_\text{text}, C_\text{traj}),
\label{ji5}
\end{equation}
where $I_0$ denotes the initial frame, $C_\text{text}$ represents the task instruction, and $C_\text{traj}$ provides sparse trajectory signals.

To bridge the gap between dense motion learning and sparse trajectory conditioning, we introduce a two-stage training strategy: 
\begin{itemize} \item \textbf{Stage 1: Dense Flow Supervision.} The model is first trained with dense optical flow, providing detailed motion cues that enable learning of accurate robot dynamics and object interactions. \item \textbf{Stage 2: Sparse Trajectory Alignment.} Training the transitions to sparse flow trajectories, aligning supervision with inference conditions and ensuring morphology-invariant motion control. 
\end{itemize} 

The two-stage design allows the generator to first acquire precise motion priors and subsequently adapt to sparse trajectory prompts, ensuring that generated videos not only follow human-demonstrated intent but also generalize across embodiments. 

\textbf{Video Policy to Robot Execution} 
The generated video $V_r$ must be mapped to executable robot actions. As illustrated in the bottom of Fig.~\ref{fig:arc}, the policy is conditioned on both the current observation and the predicted video. To incorporate $V_r$, the video frames are temporally aggregated into a compact reference image, which is subsequently projected into the model space and fused with the current state embedding. The fused representation is then decoded into a action sequence:
\begin{equation}
    A_r = F(O, S, V_r),
\end{equation}
where $O$ denotes the robot’s real-time observation, $S$ the low-level state information, $V_r$ the generated video, and $F(\cdot)$ the policy network integrating all inputs.

\subsection{Cross-embodiment Skill Transfer}  
The preceding components together establish the foundation for cross-embodiment transfer. First, the embodiment-invariant flow sampling module extracts sparse optical flow trajectories that abstract away embodiment-specific appearance and kinematics. Second, the two-stage training pipeline realizes trajectory conditioned robot execution, where sparse trajectories from human demonstrations guide the generation of robot-executable task videos, which are subsequently translated into actions through video-policy to robot execution. By combining these two components, our framework enables one-shot cross-embodiment imitation. The formulation allows a robot to reproduce human-demonstrated skills without paired datasets or reinforcement learning, bridging the embodiment gap and enabling scalable trajectory conditioned skill transfer.
\begin{table*}[ht]
\centering
\small

\setlength{\tabcolsep}{8pt} 
\renewcommand{\arraystretch}{2.1} 
\begin{tabular}{cccccccc}
\toprule
\multirow{2}{*}{Method} &\multirow{2}{*}{Publication} & \multicolumn{2}{c}{MetaWorld} & \multicolumn{2}{c}{Franka} & \multirow{2}{*}{Resolution}   & \multirow{2}{*}{Frames} \\
\cmidrule(lr){3-4} \cmidrule(lr){5-6} &
 & FVD ($\downarrow$) & KVD ($\downarrow$) & FVD ($\downarrow$) & KVD ($\downarrow$) & & \\
\midrule
AVDC (V+Lang.)      &\textit{ ICLR 2024} & 1467.68 & 1632.93 & 925.27 & 775.19 & 128$\times$128 & 8  \\
This\&That (V+Lang.) & \textit{ICRA 2025} & 857.08  & 796.31  & 991.87 & 987.20 & 448$\times$448 & 28 \\
CogVideo (V+Lang.)   & \textit{ICLR 2025 }& 528.04  & 422.86  & 427.30 & 255.82 & 480$\times$720 & 49 \\
Ours (V+Lang.+Traj)  &-------- & \textbf{318.83} & \textbf{268.03} & \textbf{309.53} & \textbf{175.01} & 480$\times$720 & 49 \\
\bottomrule
\end{tabular}
\caption{Quantitative evaluation on MetaWorld and Franka. Results are reported in terms of FVD and KVD (lower values indicate better performance). TrajSkill consistently outperforms all baselines.}
\label{tab:video_quality}
\end{table*}

\section{Experiments}

We design comprehensive experiments to evaluate the proposed {TrajSkill} framework in terms of (i) {trajectory controllability} in generated videos, (ii) {cross-embodiment transfer} from human demonstrations to robots, and (iii) {robot execution} for downstream manipulation tasks.

\subsection{Experimental Setup}

\subsubsection{Datasets}
We conduct evaluations on three representative benchmarks:
\begin{itemize}
    \item \textbf{MetaWorld 50 Tasks}~\cite{mclean2025meta}: a suite of 50 simulated manipulation tasks on a Sawyer arm, covering four difficulty levels (\textit{easy, medium, hard, very hard})~\cite{seo2023masked}.
    \item \textbf{Franka Multi-Tasks}~\cite{song2025hume}: real-world demonstrations across 14 diverse tasks with a Franka Panda robot.
    \item \textbf{XSkill}~\cite{xu2023xskill}: a hybrid dataset supporting cross-embodiment transfer, containing both sphere-agent trajectories in simulation and real human-hand demonstrations.
\end{itemize}

\subsubsection{Baselines}
We benchmark against both video generation and robot execution approaches:
\begin{itemize}
    \item \textbf{Video Generation:} 
{AVDC}~\cite{ko2023learning}, {This\&That}~\cite{wang2024language},  and {CogVideo}~\cite{yang2024cogvideox},  
    \item \textbf{Robot Execution:} {Diffusion Policy (DP)}~\cite{chi2023diffusion}, {TinyVLA}~\cite{wen2025tinyvla}, {SmolVLA}~\cite{shukor2025smolvla}, and {OCTO}~\cite{team2024octo}.
\end{itemize}
\begin{figure}[]
    \centering    \includegraphics[width=1.0\linewidth]{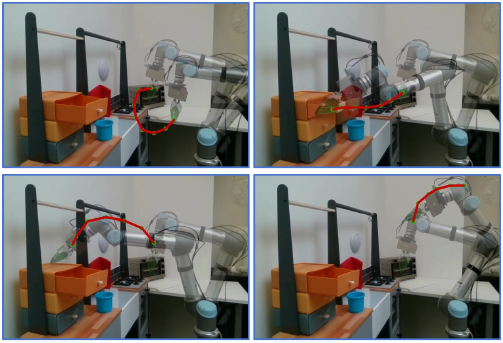}
        \caption{Trajectory conditioned video generation. The robot arm is provided with an initial frame and a predefined trajectory, shown as red curves. TrajSkill generates a sequence of motion frames where the robot follows the specified path. The figure illustrates the robotic arm at both the starting and ending points of the trajectory.}
    \label{fig:sim_fig2}
    \vspace{-13pt} 
\end{figure}

\subsubsection{Evaluation Metrics}
We evaluate the generated videos using two key metrics: Fréchet Video Distance (FVD)~\cite{unterthiner2018towards} and Kernel Video Distance (KVD)~\cite{gu2023seer}, which measure the realism and temporal consistency of the videos. Additionally, we assess the Success Rate (SR)~\cite{qu2025spatialvla}, which represents the percentage of completed manipulation tasks and serves as the ultimate metric for skill transfer.

\subsection{Trajectory Conditioned Video Generation}

\subsubsection{Visual Quality}
We evaluate on 57 test sequences sampled from MetaWorld and Franka datasets. Table~\ref{tab:video_quality} shows that {TrajSkill} achieves the best FVD and KVD across datasets, outperforming both robotics-specific VDMs and large-scale video generators. TrajSkill reduces FVD by 39.6\% and KVD by 36.6\% on MetaWorld, and by 27.6\% (FVD) and 31.6\% (KVD) on Franka compared to CogVideo. Moreover, our method produces longer and clearer videos, crucial for manipulation planning. 

\subsubsection{Trajectory Controllability}
To further evaluate the controllability of our framework, we condition the video generator on trajectories and measure whether the generated robot motions remain faithful to the given paths. As illustrated in Fig.~\ref{fig:sim_fig2}, the generated motions closely align with the input trajectories across both simple linear paths and more complex curved patterns.

From a cross-embodiment perspective, this result highlights that the 2D trajectory abstract away embodiment-specific morphology yet still convey precise spatiotemporal guidance. In practice, this means that demonstrations performed by human hands can be faithfully reinterpreted into robot-consistent motion videos, effectively bridging the gap between human and robotic embodiments.

\begin{table}[t]
  \centering
  \small
\renewcommand{\arraystretch}{1.8} 
  \begin{adjustbox}{width=1.0\linewidth} 
  \begin{tabular}{cccccc}
    \toprule
    \multirow{2}{*}{Method}         & \multicolumn{5}{c}{Success Rate (\%)} \\ 
    \cmidrule(lr){2-6}  
                    & \textit{Easy} & \textit{Medium} & \textit{Hard} & \textit{Very Hard} & Overall \\
    \midrule
    Diffusion Policy & 23.1 & 10.7   & 1.9  & 6.1       & 10.5    \\
    TinyVLA          & 77.6 & 21.5   & 11.4 & 15.8      & 31.6    \\
    SmolVLA          & 74.6 & \textbf{30.9}   & 18.0 & 30.0      & 38.3    \\ 
    Ours             &  \textbf{81.8} &  {29.1}   &  \textbf{38.0} &  \textbf{30.0}      &  \textbf{44.7}    \\ 
    \bottomrule
  \end{tabular}
  
\end{adjustbox}
\caption{Success rate comparison across task difficulties. Performance comparison of different methods on 49 robot tasks categorized by difficulty levels, showing success rates in percentage.}
\label{table1}
\end{table}

\subsection{Cross-embodiment Skill Transfer}

\subsubsection{Simulation Transfer (Sphere $\rightarrow$ Robot)}
\begin{figure*}[h]
    \centering
\includegraphics[width=1.0\textwidth]{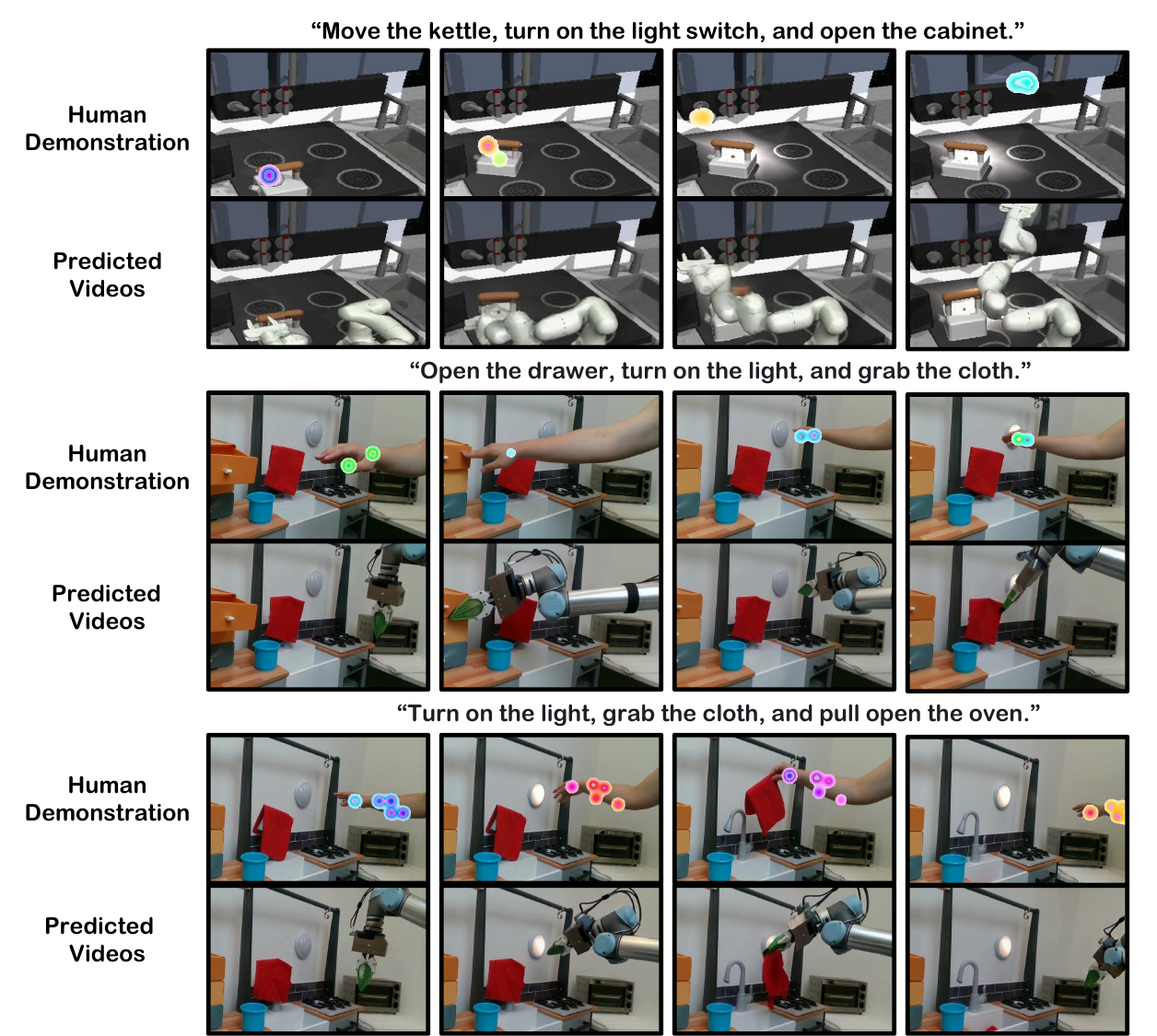}
    \caption{{Trajectory conditioned cross-embodiment skill transfer.} Top two rows show simulation results where human demonstrations are abstracted as spherical trajectories (first row) to guide robotic arm motion generation (second row). Bottom four rows demonstrate real-world transfer from human hand demonstrations to robotic arm execution for  complex multi-step tasks.}
    \label{fig:sim_fig3}
    \vspace{-13pt}  
\end{figure*}

Using XSkill’s sphere-agent demonstrations, we extract sparse flow trajectories to condition robot motion generation. As shown in Fig.~\ref{fig:sim_fig3} (top), generated videos follow the demonstrated motions, confirming accurate simulation-to-robot transfer.

\subsubsection{Real-World Transfer (Human $\rightarrow$ Robot)}
We further evaluate transfer from human hand demonstrations. Sparse trajectories extracted from human videos effectively guide robot motion generation, as shown in the middle and bottom of Fig.~\ref{fig:sim_fig3}. These results demonstrate that TrajSkill bridges embodiment gaps in both simulated and real-world scenarios.

These cross-embodiment experiments highlight TrajSkill’s ability to generalize motion knowledge across distinct agents, from simulated spheres to robots and from human hands to robotic manipulators. By demonstrating consistent trajectory transfer in both controlled and real-world settings, we establish a strong foundation for translating generated video policies into executable robot actions. Building on this, the following section investigates how such transferred skills materialize in actual robot execution, validating their practicality and robustness in complex environments.

\begin{table}[t]
\centering
\small
\renewcommand{\arraystretch}{1.5} 
\tiny 
\begin{adjustbox}{width=1.0\linewidth} 
\begin{tabular}{cccc}
\hline
Method & Octo & Diffusion Policy & Ours \\
\hline
Pick   & 0.0\%  & 81.8\% &  \textbf{90.9\%} \\
Place  & 0.0\%  &  72.7\% &  \textbf{81.8\%} \\
\hline
\end{tabular}
\end{adjustbox}
\caption{Real robot experimental results on "Put the Banana in the Basket" task. 
Success rates for Pick and Place actions across different methods.}
\label{table2}
\end{table}

\subsection{Robot Execution}

\subsubsection{Simulation Rollouts}
We evaluate robot execution of TrajSkill on the MetaWorld 50 benchmark. As reported in Table~\ref{table1}, TrajSkill achieves the highest overall success rate (44.7\%), outperforming prior approaches by a clear margin. Specifically, our method excels in both easy and hard tasks, reaching 81.8\% and 38.0\% SR respectively, while maintaining strong performance in the very hard category (30.0\%). In contrast, TinyVLA and SmolVLA perform competitively on medium tasks but drop significantly on hard tasks. Diffusion Policy shows limited generalization across all difficulty levels, with only 10.5\% overall SR. These results highlight that trajectory conditioned video generation not only enables execution but also scales effectively to more challenging scenarios, offering robust performance across varying task complexities. our method achieves competitive SR compared to Diffusion Policy, TinyVLA, and SmolVLA, verifying that trajectory conditioned video generation provides robot execution.
\begin{figure*}[!htbp]
    \centering
\includegraphics[width=1.0\linewidth]{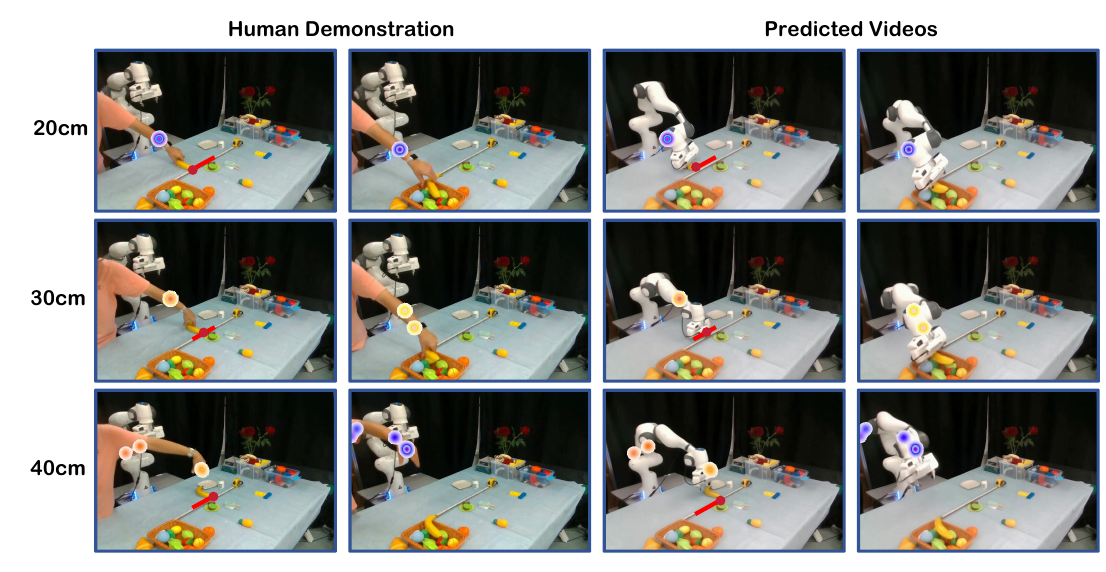}
    \caption{Human-Controlled Robot Video Prediction for Pick and Place Tasks. Human demonstrations (left) control robot arm movements in predicted videos (right) at three different banana positions: 20cm, 30cm, and 40cm from the basket.}
    \label{fig:real_fig1}
    \vspace{-4.5mm}  
\end{figure*}

\subsubsection{Real-Robot Experiments}

We deploy our TrajSkill on a Franka Panda in a kitchen environment. As illustrated in Fig.~\ref{fig:real_fig1}, a banana is placed at varying distances (20\,cm, 30\,cm, 40\,cm), and human hand videos provide sparse trajectories to guide motion generation. TrajSkill then transforms the generated video policy into executable robot actions. As shown in Table~\ref{table2}, TrajSkill achieves the highest success rates in both pick and place tasks, with 90.9\% and 81.8\% respectively, which shows that TrajSkill significantly outperforms DP and OCTO in both picking and placing success rates. These results confirm that TrajSkill not only generalizes well in simulation but also transfers effectively to real-world robot execution, achieving robust performance in challenging manipulation scenarios.

\section{Conclusions}

In this work, we introduce {TrajSkill}, a trajectory conditioned framework for cross-embodiment skill transfer. 
Our key idea is to leverage {sparse optical flow trajectories} extracted from human demonstrations as an embodiment-invariant representation of motion intent. 
Through trajectory conditioned robot execution, TrajSkill enables direct mapping from human demonstrations to robotic execution, effectively bridging the embodiment gap. Extensive experiments validate the effectiveness of the proposed framework. This work suggests a promising direction for scalable robot learning from unstructured human video demonstrations. 

Future work involves extending trajectory-based conditioning to more complex long-horizon tasks, incorporating language grounding for more detailed task specifications, and applying the framework to a variety of robot morphologies in open-world environments.

\bibliographystyle{IEEEtran}
\bibliography{reference}

\end{document}